%% file: main.tex
\documentclass[10pt,twocolumn,letterpaper]{article}
\usepackage{iccv}
\usepackage{times}
\usepackage{amsmath}
\usepackage{amssymb}
\usepackage{bm}
\usepackage{graphicx}
\usepackage{todonotes}
\usepackage{pdfpages}
\usepackage[pagebackref=true,breaklinks=true,letterpaper=true,colorlinks,bookmarks=false]{hyperref}
\usepackage{cleveref}
\usepackage{booktabs}
\usepackage{comment}
\usepackage{algorithm}
\usepackage{algorithmicx}
\usepackage[noend]{algpseudocode}

\presetkeys{todonotes}{inline,backgroundcolor=yellow}{}

\newcommand{\bbm}{\bm{m}}

\newcommand{\bw}{\bm{w}}
\newcommand{\bx}{\bm{x}}

\iccvfinalcopy
\def\iccvPaperID{1968}
\ificcvfinal\fi

\title{NormGrad: Finding the Pixels that Matter for Training}

\author{
Sylvestre-Alvise Rebuffi$^{1}$ \quad
Ruth Fong$^{1}$ \quad
Xu Ji$^{1}$ \quad
Hakan Bilen$^{2}$ \quad
Andrea Vedaldi$^{1}$ \vspace{1em}\\
\centering
\begin{minipage}{.4\textwidth}
\centering
$^1$\small{Visual Geometry Group\\ University of Oxford\\}
{\tt\small srebuffi@robots.ox.ac.uk}
\end{minipage} 
\begin{minipage}{.4\textwidth}
\centering
$^2$\small{School of Informatics\\University of Edinburgh\\}
\end{minipage}
}

\begin{document}
\maketitle
\begin{abstract}
The different families of saliency methods, either based on contrastive signals, closed-form formulas mixing gradients with activations or on perturbation masks, all focus on which parts of an image are responsible for the model's inference. In this paper, we are rather interested by the locations of an image that contribute to the model's training. First, we propose a principled attribution method that we extract from the summation formula used to compute the gradient of the weights for a $1 \times 1$ convolutional layer. The resulting formula is fast to compute and can used throughout the network, allowing us to efficiently produce fined-grained importance maps. We will show how to extend it in order to compute saliency maps at any targeted point within the network. Secondly, to make the attribution really specific to the training of the model, we introduce a meta-learning approach for saliency methods by considering an inner optimisation step within the loss. This way, we do not aim at identifying the parts of an image that contribute to the model's output but rather the locations that are responsible for the good training of the model on this image. Conversely, we also show that a similar meta-learning approach can be used to extract the adversarial locations which can lead to the degradation of the model.

\end{abstract}

\input{method}

\input{experiments}

\input{conclusions}

\paragraph{Acknowledgments:} This work acknowledges the support of Mathworks/DTA DFR02620, EPSRC SeeBiByte and ERC 677195-IDIU.

\bibliographystyle{plain}
\bibliography{refs}

\end{document}

%% file: method.tex
\section{Method}\label{s:method}

Consider a training set $\mathcal{D}$ of pairs $(\bx,y)$ where $\bx \in \mathbb{R}^{H\times W\times 3}$ are (color) images and $y \in \{1,\dots,C\}$ their labels.
Furthermore, let $y = \Phi_\theta(\bx)$ be a model such as a deep neural network whose parameters $\theta$ are optimized using the cross-entropy loss $\ell$ to predict labels from images.
We are interested in finding which regions in these images are important for accurately training the parameters $\theta$ of the model.
We propose two approaches to do so.

\subsection{NormGrad}\label{s:gradmethod}

We propose a method based on the fact that the gradients of the weights from convolutional layers come from the summing over spatial locations of products between activation gradients and features. By considering these products before the spatial summing, we can extract a gradient importance map. As we take the Frobenius norm of these local products, we dub our method "NormGrad".

In order to do so, we focus our attention on a targeted convolutional layer in the network.
Assuming that the network is a simple chain (other topologies are treated in the same manner but the notation is more complex), we can write $\ell \circ \Phi = h \circ k_{\bw} \circ q$ where $k_{\bw}$ is the convolutional layer, $h$ the composition of all layers that follow that (including the loss) and $q$ the composition of layers that precede it.
If $\bx' = q(\bx) \in \mathbb{R}^{C'\times H'\times W'}$ is the input to the convolutional layer, the output $\bx'' = k_{\bw}(\bx')$ is the convolution $\bx'' = \bw \ast \bx'$ where tensor $\bw \in \mathbb{R}^{C''\times C'\times H_k \times W_k}$ contains the parameters of the filter bank (we ignore the biases as they are immaterial).

As the notation is rather cumbersome, we first explain the easier case in which the filter bank has spatial dimensions $H_k=W_k=1$ and then sketch the general case.
In the simple case, the output elements of convolution are given, for all $v=1,\dots,H',~u=1,\dots,H'$ by the expression
$$
x''_{kvu}
=
\sum_{c=1}^{C'} w_{kc} x'_{cvu},
$$
Computing the gradient with respect to the filter bank parameters yields:
\begin{equation*}
\frac{dh(\bw \ast \bx')}{dw_{pq}}
=
\sum_{k=1}^{C''}
\sum_{u\in\Omega''}
\frac{dh}{dx''_{ku}}
\frac{dx''_{ku}}{dw_{pq}}
=
\sum_{u\in\Omega''}
\frac{dh}{dx''_{qu}}
x'_{pu}.
\end{equation*}
We can summarize this in matrix form as
\begin{equation}
\frac{dh(\bw \ast \bx')}{d\bw}
=
\sum_{u\in\Omega''}
\bm{g}_{u} {\bx'}_{u}^\top.
\end{equation}
Note that each spatial location $u$ contributes to the gradient via the outer product of $\bm{g}_u$ (containing column $u$ of the backpropagated gradient) and $\bx'_u$ (containing column $u$ of the intermediate activation tensor $\bx$).
We thus generate an importance map by computing the Frobenius norm of each of these products.
Since
$
\|\bm{g}_{u} {\bx'}_{u}^\top\|_F 
=
\|{\bx'}_{u}\| \cdot
\|\bm{g}_{u}\|,
$
the importance map is
$$
\bbm_{u} = 
\|{\bx'}_{u}\| \cdot
\|\bm{g}_{u}\|.
$$

\paragraph{General filter shapes.}

For filters whose spatial dimensions is not $1\times 1$, we can rewrite convolution using the matrix form:
$$
  X'' = \iota(\bx') W,
$$
where $X'' \in \mathbb{R}^{C'' \times (H''W'')}$ and $W \in \mathbb{R}^{C''\times (C'H_kW_k)}$ are the output and filter tensor reshaped as matrices and $\iota(\bx) \in \mathbb{R}^{(H''W'') \times (C'H'W')}$ is a matrix whose rows contain the patches of the input tensor to which the filters are applied (this operation is often called \texttt{im2row}).

With slight abuse of notation, as the network is trained, the filter weights $W$ are updated according to the gradient
$$
\frac{dh(\iota(\bx') W)}{dW}
=
\frac{dh}{dX''}
\cdot 
\frac{dX''}{dW}
=
\frac{dh}{dX''}
\cdot
\iota(\bx').
$$
This expression shows how the output of the network depends on the interaction of certain patches in the intermediate tensor $\bx'$ and corresponding filter weights.
We compute the product of the norm of such patches and weights as before, and accumulate them to each pixel contained in the patch to obtain the corresponding mask values.

\paragraph{ 1$\times$1 identity convolution trick.} Due to the facts that the computation for general filter shapes can be heavy and that we would like to compute saliency maps at any point of the network (and not necessarily at convolutional layers), we propose the trick of considering at the targeted point of the network an imaginary 1$\times$1 convolution whose weights are initialised with identity. In this case, we have $\bx' = \bx''$ and our importance map can be written:
$$
\bbm_{u}	=  \|\bm{g}_{u} {\bx''}_{u}^\top\|_F. 
$$

\paragraph{Comparison to Grad-CAM.}
With this previous trick we can directly compare our method with Grad-CAM~\cite{selvaraju17grad-cam:} as they use the same $\bm{g}$ and $\bx''$. Grad-CAM can be written in our notation:
$$
\bbm_{u}	=  \max (\bar{\bm{g}}^\top {\bx''}_{u}, 0).
$$
where $ \bar{\bm{g}} $ is the mean of the gradient over the spatial locations. Hence each pixel's importance depends on the gradient at the other spatial locations whereas our formulation treats each pixel independently. If we consider the input of a global average pooling layer like ResNet, the gradient is uniform over spatial locations and Grad-CAM (GC) can be written as a function of our method NormGrad (NG):
$$
\bbm^{GC}_{u}	=  \cos(\bm{g}_{u}, {\bx''}_{u})_+ \cdot \bbm^{NG}_{u}.
$$
In this case, Grad-CAM can be interpreted as putting on top of our method a filter that only lets through the locations where the gradients and features are positively aligned. As our method highlights the locations that matter for training, it also entails the locations corresponding to adversarial classes. Therefore our method is not as class-selective as Grad-CAM (\Cref{fig:ablation}).

Finally, Grad-CAM implicitly requires the use of the 1$\times$1 identity convolution trick and could not be used directly on a convolutional layer with different input and output channel dimensions as it would not be possible to compute the inner product between the input features and the activation gradients which would not have the same dimensions. Our method does not suffer from this drawback as it relies on the norm of the outer product of the features and the gradients.

\subsection{NormGrad at order 1}\label{s:order1}
As such, our method produces the same importance map whether we would like to maximize or minimize the loss $\ell$. Hence, the same locations will be highlighted whether we want to train on the image or to minimize the model's performance. Though, it is possible to make the importance map specific to the training process by backpropagating from an order 1 loss. To do so, we want to minimize:
\begin{equation}
  \ell(\theta - \epsilon \nabla_\theta \ell(\theta, x), x) .
\end{equation}
where $\theta'= \theta - \epsilon \nabla_\theta \ell(\theta, x)$ corresponds to one step of SGD of learning rate $\epsilon$. Using an optimisation step within a loss has been commonly used in meta-learning~\cite{finn2017model} or architecture search~\cite{liu2018darts}.The gradient of this loss is:
\begin{equation*}
 \nabla_{\theta'} \ell(\theta', x) - \epsilon \nabla^2_\theta \ell(\theta, x) \nabla_{\theta'} \ell(\theta', x).
\end{equation*}
Using a centered finite difference scheme of step $h$ as in~\cite{Pearlmutter94fastexact}, we can approximate the hessian-vector product by:
\begin{equation*}
\nabla^2_\theta \ell(\theta, x) \nabla_{\theta'} \ell(\theta', x) = \frac{\nabla_{\theta^+} \ell(\theta^+, x) - \nabla_{\theta^-} \ell(\theta^-, x)}{2h} +O(h).
\end{equation*}
where $\theta^\pm = \theta \pm h \nabla_{\theta'} \ell(\theta', x)$. As above, the local contribution to the gradient of the weights can be written as a product of the input features and the activation gradients. Thus, if we note $\bm{g}_{u, \theta'}$, the column $u$ of the backpropagated gradient when using the parameters $\theta'$, we get the following importance map: 
$$
\bbm_{u}	=  \|\bm{g}_{u, \theta'} {\bx'}_{u, \theta'}^\top - \frac{\epsilon}{2h} (\bm{g}_{u, \theta^+} {\bx'}_{u, \theta^+}^\top - \bm{g}_{u, \theta^-} {\bx'}_{u, \theta^-}^\top) \|_F. 
$$
Empirically, we notice that for the activation maps $x'$, we have $ x'_{\theta'} \approx x'_{\theta^\pm} $. So we can factorize the sum of products by the activation $ x'_{\theta'}$ and use again the property that the Frobenius norm of the outer product of two vectors is the product of the norms of the vectors to get the importance map:
$$
\bbm_{u} \approx
\|\bm{g}_{u, \theta'} - \frac{\epsilon}{2h} (\bm{g}_{u, \theta^+} - \bm{g}_{u, \theta^-} )\| \cdot
\|{\bx'}_{u, \theta'}\|.
$$
We notice that if we take $\epsilon \to 0$, this formula boils back down to the formula of our method at order 0. In practice, as for order 0, we can use the 1$\times$1 identity convolution trick and thus in the formula $\|{\bx'}_{u, \theta'}\|$ gets replaced by $\|{\bx''}_{u, \theta'}\|$.

Conversely, if we would like to get an importance map which would highlight the sensitive pixels for the degradation of the model's performance, we should minimize the loss $-\ell(\theta + \epsilon \nabla_\theta \ell(\theta, x), x)$ and we get the following adversarial importance map:
$$
\bbm_{u}^{adv} \approx
\|\bm{g}_{u, \theta'} + \frac{\epsilon}{2h} (\bm{g}_{u, \theta^+} - \bm{g}_{u, \theta^-} )\| \cdot
\|{\bx'}_{u, \theta'}\|.
$$
where $\theta'= \theta + \epsilon \nabla_\theta \ell(\theta, x)$.

%% file: experiments.tex
\section{Experiments}\label{s:experiments}

\begin{figure}[!htb]
    \centering
    \includegraphics[width=\linewidth]{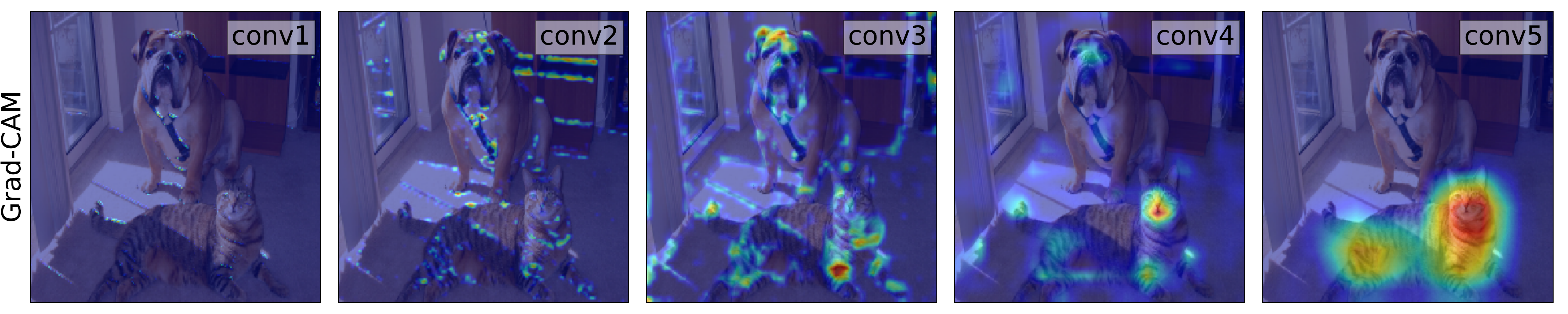}
    \includegraphics[width=\linewidth]{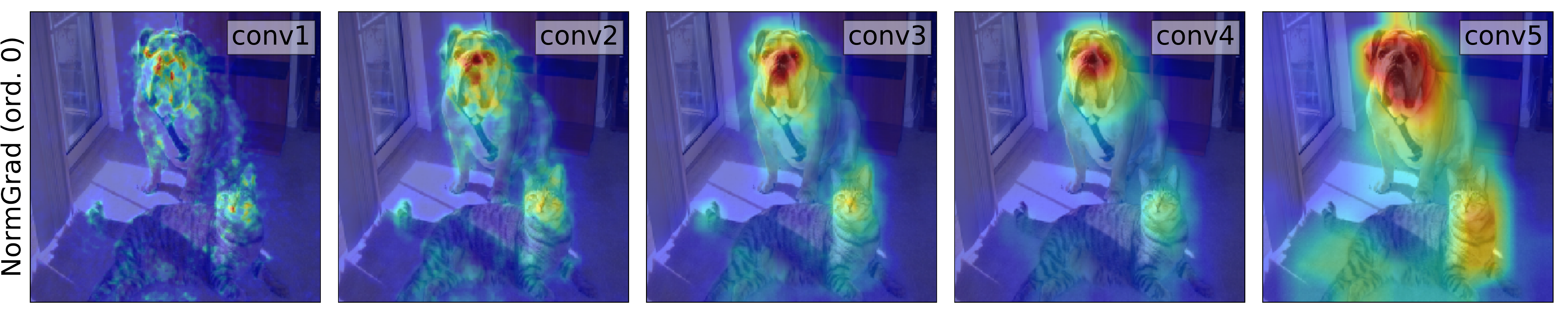}
    \includegraphics[width=\linewidth]{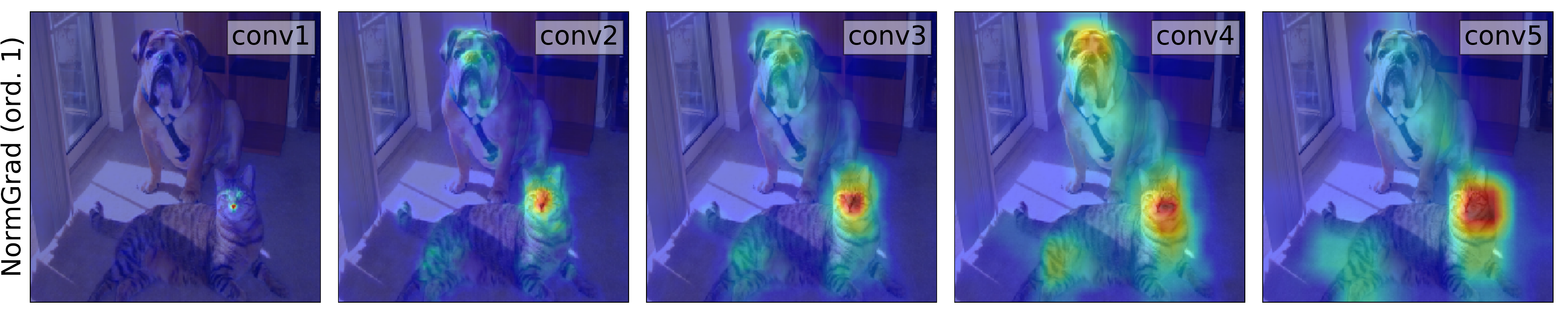}
    \includegraphics[width=\linewidth]{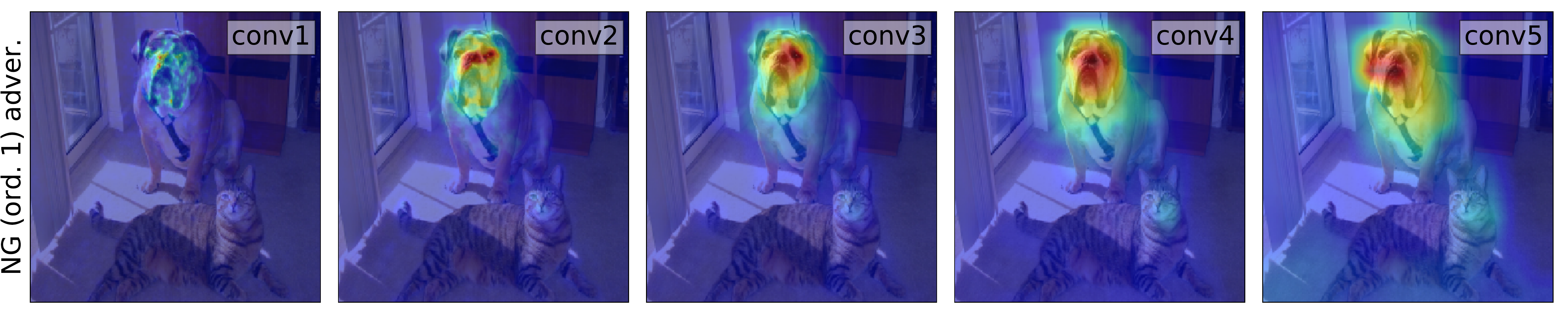}
    \caption{Grad-CAM vs. NormGrad targeting the "tiger cat" class at different depths of VGG-16. Grad-CAM only works at the end of the network i.e. last two pictures of the first row. In the second row, our method at order 0 provides masks at any layer but is not class-selective. This problem is alleviated by using the order 1 method as seen in the third row. Finally, the last row is the adversarial case where we use $\bbm_{u}^{adv}$ to highlight the pixels responsible for the degradation of the model performance (the dog in our case). } 
    \label{fig:ablation}
\end{figure}

\begin{figure}[!htb]
    \centering
    \includegraphics[width=\linewidth]{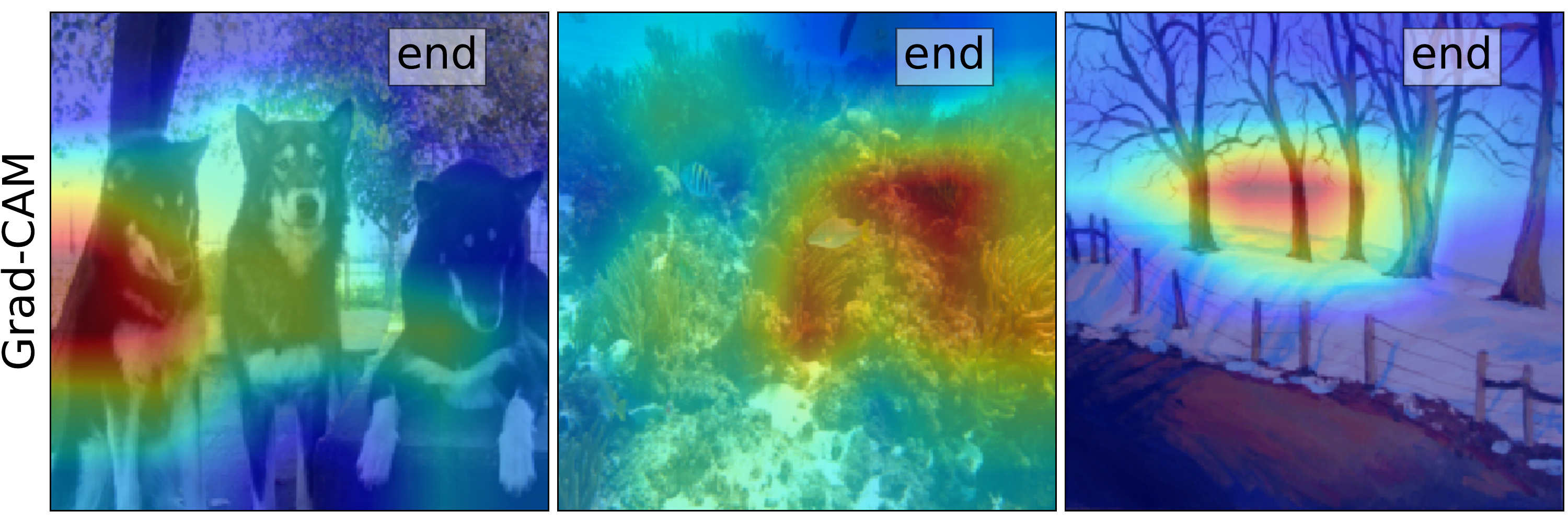}
    \includegraphics[width=\linewidth]{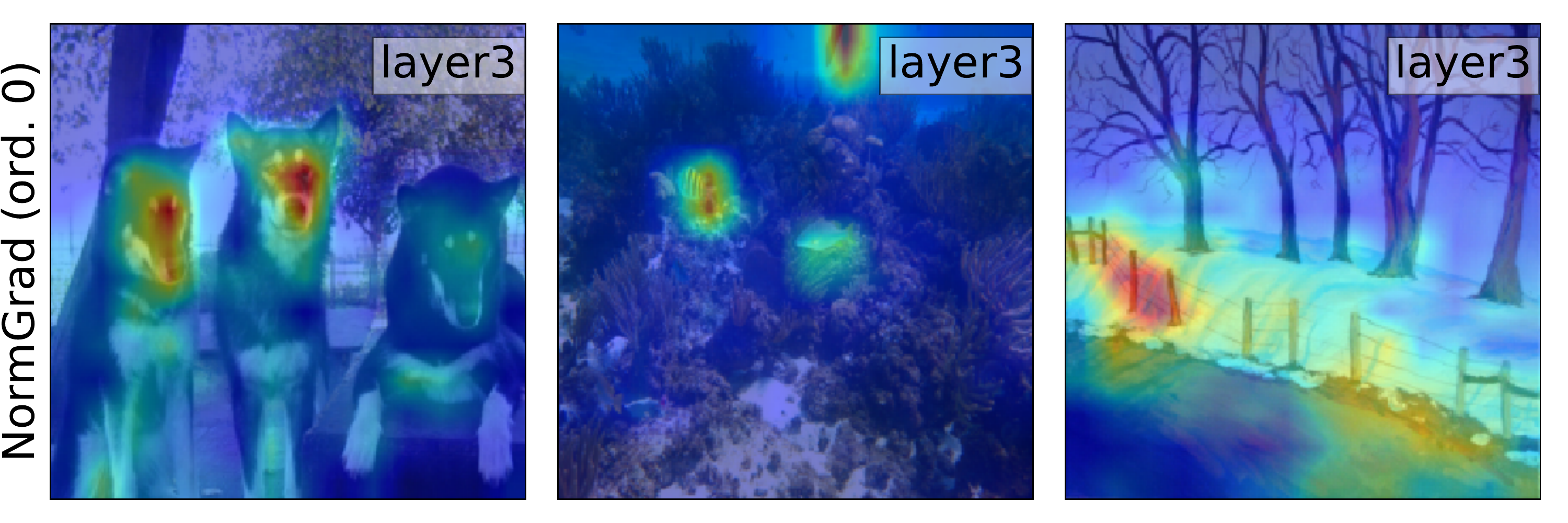}
    \includegraphics[width=\linewidth]{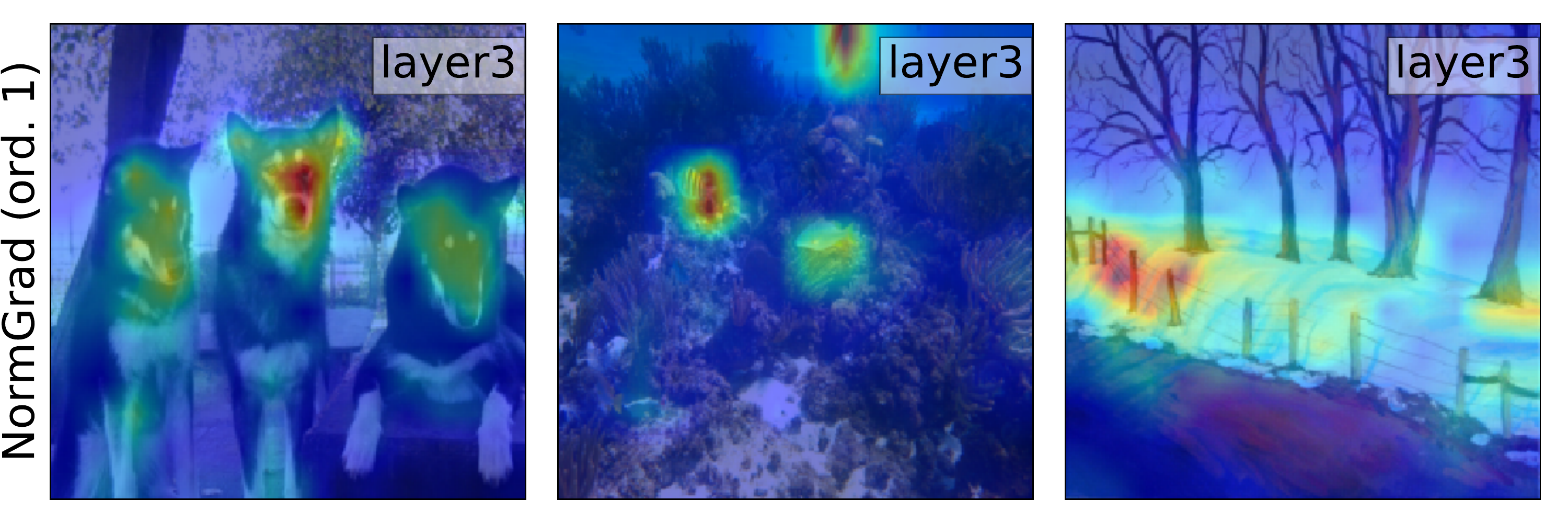}
    \caption{Grad-CAM vs. NormGrad targeting respectively the classes "malamute", "coral reef" and "fence" on ResNet-50. For our method (last two rows) we show the importance maps after the first 3$\times$3 convolution in the third macro-block. Contrary to Grad-CAM, our method provides high resolution maps highlighting fine grained details like the head of the dogs in the first image or fishes in the second one.} 
    \label{fig:resnet}
\end{figure}

\begin{figure}[!htb]
    \centering
    \includegraphics[width=\linewidth]{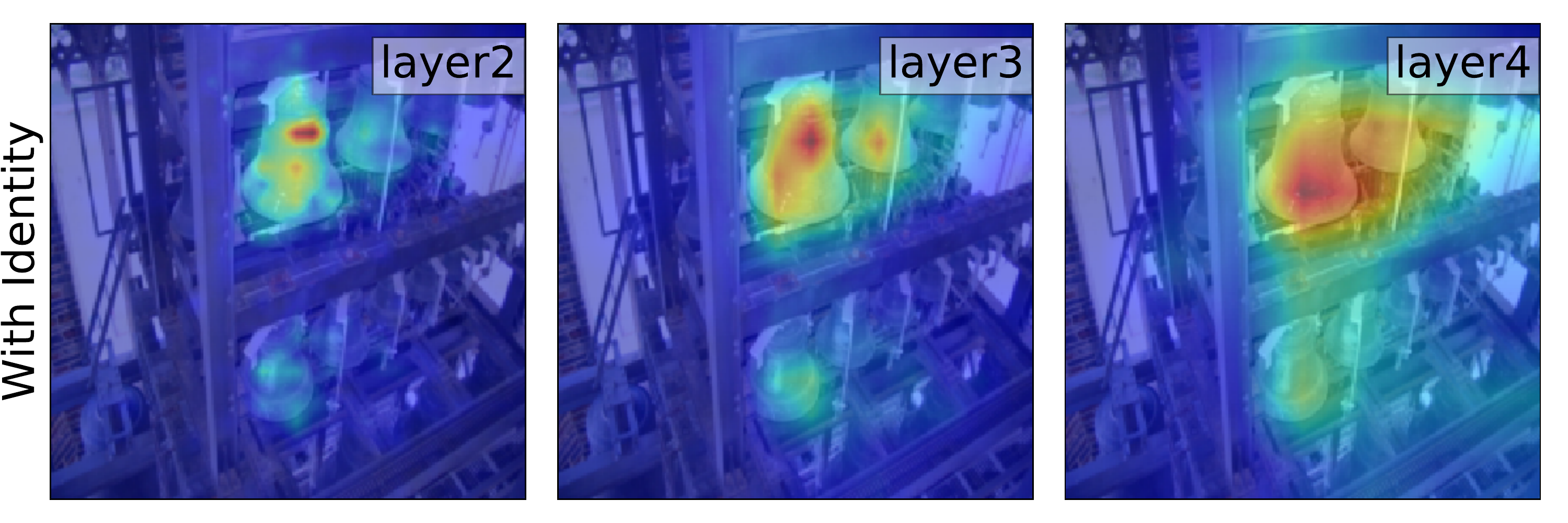}
    \includegraphics[width=\linewidth]{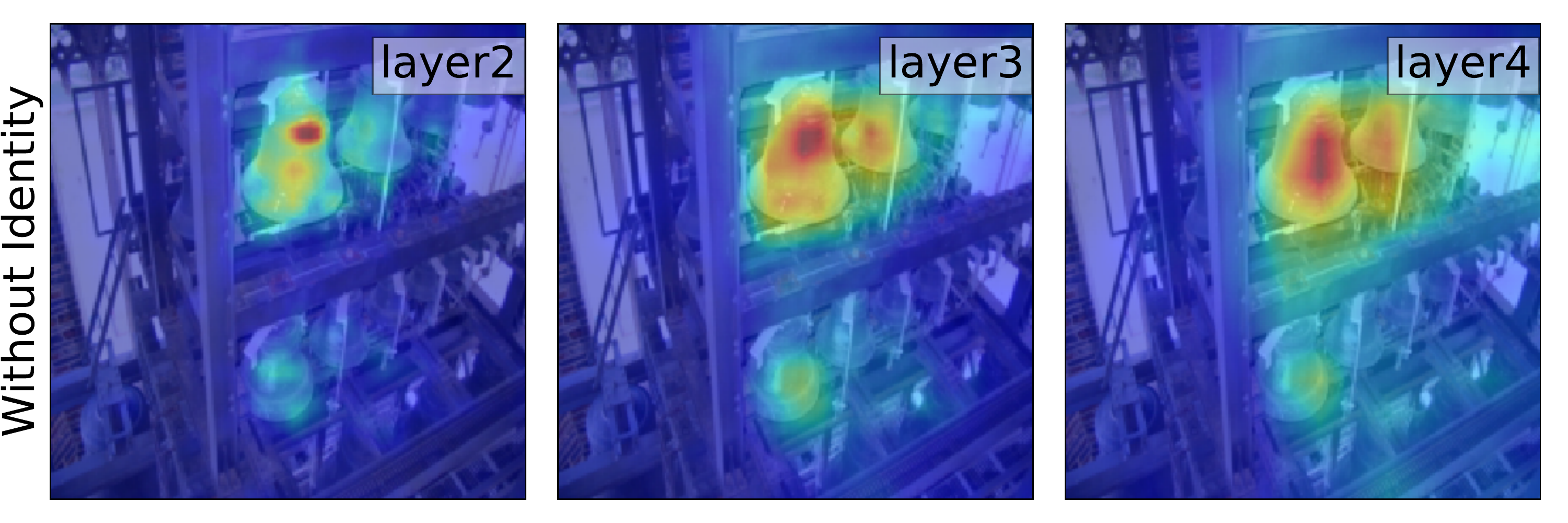}
    \caption{Effect of the 1$\times$1 identity convolution trick on a "bell" image with ResNet-50. We do a comparison on the 1$\times$1 downsample convolution of the second, third and fourth macro-blocks between considering a 1$\times$1 identity convolution after the downsample convolution (first row) and directly analyzing the downsample convolution without added identity (second row). We notice that both approaches produce similar masks.} 
    \label{fig:id_trick}
\end{figure}

\paragraph{Experimental set-up.} 
We evaluate our proposed saliency method on images from ImageNet~\cite{russakovsky2015imagenet} with resolution 224$\times$224. We compare it with the standard saliency method Grad-CAM on different architectures like VGG-16~\cite{simonyan14deep} or ResNet-50~\cite{he2016deep}. The order 0 version of our method and Grad-CAM do not have any hyperparameters to tune and are simply computed with their respective formulae. For our method at order 1, we take the same hyperparameters for all settings: $\epsilon = 0.0005$ for the learning rate of the inner SGD step and $h = 0.5 / \|\nabla_{\theta'} \ell(\theta', x)\|_2$ for the finite difference step used in the hessian-vector product approximation.

\paragraph{Computing saliency at different depths.}
In Figure~\ref{fig:ablation}, we apply Grad-CAM and our proposed gradient-based method at 5 intermediate points within VGG-16; specifically, at the end of each "conv" block, in order to test these saliency methods at different network depths and spatial resolutions. We notice that when Grad-CAM can produce significant masks only at the last two points of the network, our method at order 0 and 1 provide interpretable masks at any of the considered points. We further observe that our order 0 method by focusing on the locations contributing the most to the gradients of the weights is not class selective. Indeed the targeted class (the cat in our case) and the adversarial class (the dog here) are highlighted as they both contribute positively and negatively to the gradient of the weights. By considering an inner optimisation step within the loss in our order 1 method, we made the saliency map specific to either training as seen of the third row of Figure~\ref{fig:ablation} or to the perturbation of the model as shown by the fourth row.

\paragraph{Efficiently producing fined-grained saliency maps.}
Like Grad-CAM, our method at order 0 requires only one forward-backward pass, is efficient as it just needs to compute the norm of vectors and can directly process minibatches of images. Hence it is even possible to compute these saliency maps during training. Our method at order 1 is more computationally expensive as 4 forward-backward passes are needed to produce the saliency map. In Figure~\ref{fig:resnet}, we apply Grad-CAM just before the last pooling layer of ResNet-50 and our method after the first $3 \times 3$ convolutional layer of the third macro-block of the same ResNet. Compared to VGG-16 in Figure~\ref{fig:ablation}, we notice that Grad-CAM produces more coarse saliency maps due to the low spatial resolution at the end of ResNet whereas our method can still produce high resolution maps. Other methods like the perturbation masks from~\cite{fong17interpretable} also provide fine-grained saliency maps but they require several hundreds of forward-backward passes when our most expensive closed-form method only requires 4 of them. Finally, we observe on the three images of Figure~\ref{fig:resnet} that when there is not a strong adversarial class as in Figure~\ref{fig:ablation} with the cat and the dog, our method produces similar masks at order 0 and 1. Therefore, when an image does not have elements from different classes, it is sufficient to just compute the saliency using the order 0 version of our method.

%% file: conclusions.tex
\section{Conclusions}\label{s:conclusions}

In this paper, we have introduced a method based on the norm of the local gradients components which, when they are summed, produce the gradient of the weights for a $1 \times 1$ convolutional layer. This results in a simple closed-form formula, easy to compute at training or testing time and that can be applied throughout the network to produce fine-grained attribution maps. We then showed that it is possible to use this method at every point of the network by considering a $1 \times 1$ identity convolution at the targeted location. Finally, we extended our method at the order 1 using a meta-learning approach to make the saliency mask specific to the training process.